\pdfoutput=1

\documentclass[11pt]{article}

\usepackage[final]{acl}

\usepackage{times}
\usepackage{latexsym}

\usepackage[T1]{fontenc}

\usepackage[utf8]{inputenc}

\usepackage{microtype}

\usepackage{inconsolata}

\usepackage{booktabs}
\usepackage{amsmath}
\usepackage{float}
\usepackage{adjustbox}
\usepackage{longtable}
\usepackage{xcolor,colortbl}
\usepackage{diagbox}
\usepackage{array,multirow}
\usepackage{graphicx} 
\usepackage{placeins} 
\usepackage{microtype} 
\usepackage{enumitem} 
\usepackage{amsmath, bm} 
\usepackage{comment} 
\usepackage{amsmath, bm} 

\title{How to Discern Important Urgent News?}

\author{Oleg Vasilyev \& John Bohannon \\
  Primer Technologies Inc. \\
  San Francisco, California \\
\texttt{\{oleg,john\}@primer.ai}\\}

\begin{document}
\maketitle
\begin{abstract}
We found that a simple property of clusters in a clustered dataset of news correlate strongly with importance and urgency of news (IUN) as assessed by LLM. We verified our finding across different news datasets, dataset sizes, clustering algorithms and embeddings. The found correlation should allow using clustering (as an alternative to LLM) for identifying the most important urgent news, or for filtering out unimportant articles.
\end{abstract}

\section{Introduction}
\label{sec:intro}
Clustering of news articles provides efficient aggregation that can be used for trend monitoring, comprehensive coverage, news categorization, and can be beneficial for content summarization. In this paper we suggest that clustering can also be used for a speedy ranking of importance and urgency of news (IUN), at least the IUN as assessed by LLM. 

Proficiency of Large Language Models (LLMs) have advanced so much that LLM may be used for annotating texts with arguable success \cite{Gilardi2023, kuzman2023chatgpt, Petter2023chatgpt4}, at least for tasks with low requirements of domain specific knowledge \cite{weber2023evaluation, lu2023human}. For goal of this paper we consider IUN as a sentiment to be scored on Likert scale. Existing related work includes, for example, classifying tweets relevance \cite{Gilardi2023} or scoring financial sentiment \cite{fatemi2023comparative}. Using LLM (or even smaller models) on large amount of daily news is not always desirable, for cost and processing time reasons. We suggest that simple properties of clusters can be a good substitute.

{\bf{Our contribution}}: We reveal and analyze strong correlation between simple properties of clustered news and IUN as assessed by GPT-3.5-Turbo. We observe the correlations on the data from four different news datasets, four different data sizes, three different clustering algorithms, and three text embeddings. We also review the correlations with IUN assessed by small classification models.

\section{Data}
\label{sec:data}
\subsection{Scoring IUN}
\label{ssec:data_scoring}
Scoring IUN of a text by LLM was done by prompting LLM to output a single token - the IUN score, defined by Likert scale 1 to 5:

\hspace{1pt}1: The text is not a news article.

\hspace{1pt}2: The news in the text can be perceived as not important and not urgent.

\hspace{1pt}3: The news in the text can be perceived as having low importance and low urgency.

\hspace{1pt}4: The news in the text can be perceived as important and urgent.

\hspace{1pt}5: The news in the text can be perceived as highly important and urgent.

For the exact prompt and more detail see Appendix ~\ref{apx:ssec:prompt}.

For comparison throughout this work we also considered a much lighter scoring alternatives to GPT-3.5: classification by {\it{bart-large-mnli}}\footnote{https://huggingface.co/facebook/bart-large-mnli} \cite{yin-etal-2019-benchmarking} and by its even lighter distilled version 
{\it{distilbart-mnli-12-1}}\footnote{https://huggingface.co/valhalla/distilbart-mnli-12-1}. 
IUN score from these models is picked up as a logit of the first class from the classification on two classes: 'urgent' and 'not urgent' (the urgency may be the main property of important urgent news, see Appendix ~\ref{apx:ssec:classif}). The scores obtained this way correlate well with the LLM generated score; the correlations are shown in Table ~\ref{tab:corr_llM_other}. 

\begin{table}[th!]
\centering
\begin{tabular}{@{}llccccc@{}}
\hline
{correlation}&{$XS$}&{$CNN$}&{$DM$}&{$WN$}\\
\hline
    {LLM-B}&{0.21}&{0.26}&{0.20}&{0.25}\\
    {LLM-D}&{0.25}&{0.22}&{0.24}&{0.20}\\
    {B-D}&{0.33}&{0.39}&{0.32}&{0.35}\\
\hline
\end{tabular}
\caption{Kendall's $\tau$ correlations between IUN scores by {\it{LLM}}, {\it{B}} and {\it{D}}. For example, the first row {\it{LLM-B}} is the correlation of IUN by {\it{LLM}} with IUN by {\it{B}}.
Columns: the datasets {\it{XS}} (XSum), {\it{CNN}}, {\it{DM}} (Daily Mail), {\it{WN}} (WikiNews).}
\label{tab:corr_llM_other}
\end{table}

In this and other tables in the paper, the scores produced by {\it{GPT-3.5-Turbo}}, by {\it{bart-large-mnli}} and by {\it{distilbart-mnli-12-1}} are denoted simply as {\it{LLM}}, {\it{B}} and {\it{D}} correspondingly. The news datasets and their notations in tables are:

\hspace{1pt}{\it{XS}}: First 20K texts of XSum dataset\footnote{https://huggingface.co/datasets/EdinburghNLP/xsum} \cite{narayan-etal-2018-dont}

\hspace{1pt}{\it{CNN}}: First 20K texts of CNN part of CNN / Daily Mail dataset\footnote{https://huggingface.co/datasets/cnn\_dailymail} \cite{cnndm, see-etal-2017-get}

\hspace{1pt}{\it{DM}}: First 20K texts of Daily Mail part of CNN / Daily Mail dataset.

\hspace{1pt}{\it{WN}}: First 20K English texts of WikiNews dataset\footnote{https://github.com/PrimerAI/primer-research} \cite{vasilyev2023linear}

Table ~\ref{tab:corr_llM_other} shows correlations of the scores obtained from the top chunk of the text; the chunk consists of as many sentences as fits into 1000 characters. Examples of texts corresponding to different IUN scores are given in Appendix ~\ref{apx:ssec:text_examples}.

\subsection{Clustering}
\label{ssec:data_clustering}

For clustering, we use the same four datasets as for the scoring, but each dataset is taken in four sizes - the number of first texts taken for clustering: $5K$, $10K$, $15K$ and all $20K$. The clustering is done on embeddings of the top chunk of each text. We consider three kinds of embeddings:

\hspace{1pt}{\it{MPN}}: all-mpnet-base-v2\footnote{https://www.sbert.net/docs/pretrained\_models.html} \cite{reimers-gurevych-2019-sentence}

\hspace{1pt}{\it{L6}}: all-MiniLM-L6-v2 (a lighter sbert embedding)

\hspace{1pt}{\it{E5}}: multilingual-e5-small\footnote{https://huggingface.co/intfloat/multilingual-e5-small} \cite{wang2022text}

Before the clustering, embeddings' dimension is reduced to $20$ by UMAP\footnote{https://github.com/lmcinnes/umap} \cite{mcinnes2020umap} with number of neighbors $10$ amd min distance $0$. Other choices of UMAP (or no UMAP applied) lead to observations similar to the ones prsented and discussed here (see Appendix ~\ref{apx:other_umap}).

We use three very different clustering algorithms: 
HDBSCAN\footnote{https://github.com/scikit-learn-contrib/hdbscan} (HDBSCAN-flat, min cluster size $5$) \cite{10.1007/978-3-642-37456-2-14, 8215642}, Agglomerative (linkage 'ward' \cite{doi:10.1080/01621459.1963.10500845}) and KMeans\footnote{https://scikit-learn.org/stable/modules/clustering.html} ('lloyd') (Appendix ~\ref{apx:clustering_algos}). Clustering is done with each algorithm targeting a number of clusters in range between 20 and 100 with step 10. 
 For our analysis we keep the clustering cases that succeeded providing not less that 20 clusters (almost all the cases), and we consider IUN for all clusters that contain at least three samples.

\begin{figure}[]
\includegraphics[width=0.48\textwidth]{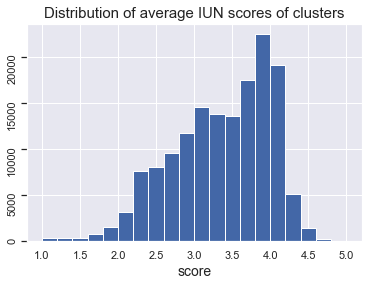}
\caption{Histogram (20 bins) of LLM generated IUN average cluster scores for clusters from all clustering cases considered in this work (see Section ~\ref{ssec:data_clustering}).}
\label{fig:Scores_All}
\end{figure}

To each cluster we can assign its IUN score as average of IUN scores of all the texts of the cluster. The distribution of the cluster LLM-generated IUN scores over all our clustering cases is shown in Figure ~\ref{fig:Scores_All}. Each cluster's score is well defined: the standard deviation over the cluster's items is typically less than $1.0$ (Appendix ~\ref{apx:ssec:histo_scores_dev}). As may be expected for a news data, most clusters have IUN around 4 and 3. Split of the histogram by datasets, dataset sizes, embeddings, clustering algorithms or UMAP versions also appear reasonable (Appendix ~\ref{apx:ssec:histo_scores_split}).

\section{IUN by Clustering vs Scores}
\label{sec:observations}

\begin{table*}[th!]
\centering
\begin{tabular}{@{}lllllllllll@{}}
\toprule
\multicolumn{2}{r}{} & \multicolumn{3}{c}{LLM} & \multicolumn{3}{c}{B} &
\multicolumn{3}{c}{D}\\
{}&{selection}&{MPN}&{L6}&{E5}&{MPN}&{L6}&{E5}&{MPN}&{L6}&{E5}\\
\hline
    \parbox[t]{2mm}{\multirow{4}{*}{\rotatebox[origin=c]{90}{dataset}}}
      &{XS}&{0.45}&{0.55}&{0.41}&{0.26}&{0.43}&{0.28}&{0.37}&{0.46}&{0.44}\\
    {}&{CNN}&{0.34}&{0.25}&{0.12}&{0.05}&{0.06}&{-0.06}&{0.26}&{0.19}&{0.11}\\
    {}&{DM}&{0.25}&{0.40}&{0.30}&{0.07}&{0.27}&{0.12}&{0.06}&{0.24}&{0.20}\\
    {}&{WN}&{0.36}&{0.36}&{0.51}&{0.23}&{0.26}&{0.38}&{0.33}&{0.39}&{0.45}\\
\hline
    \parbox[t]{2mm}{\multirow{4}{*}{\rotatebox[origin=c]{90}{data size}}}
      &{5000}&{0.36}&{0.39}&{0.31}&{0.15}&{0.25}&{0.16}&{0.29}&{0.34}&{0.29}\\
    {}&{10000}&{0.36}&{0.40}&{0.34}&{0.15}&{0.27}&{0.18}&{0.25}&{0.34}&{0.30}\\
    {}&{15000}&{0.34}&{0.38}&{0.33}&{0.15}&{0.26}&{0.20}&{0.26}&{0.33}&{0.32}\\
    {}&{20000}&{0.36}&{0.39}&{0.34}&{0.16}&{0.25}&{0.18}&{0.24}&{0.29}&{0.29}\\
\hline
    \parbox[t]{2mm}{\multirow{3}{*}{\rotatebox[origin=c]{90}{clusters}}}
      &{HDBSCAN}&{0.43}&{0.46}&{0.39}&{0.20}&{0.29}&{0.25}&{0.29}&{0.34}&{0.34}\\
    {}&{KMeans}&{0.31}&{0.36}&{0.31}&{0.13}&{0.24}&{0.15}&{0.24}&{0.31}&{0.29}\\
    {}&{Agglomerative}&{0.32}&{0.36}&{0.30}&{0.13}&{0.24}&{0.15}&{0.24}&{0.32}&{0.28}\\
\hline
    \parbox[t]{2mm}{\multirow{3}{*}{\rotatebox[origin=c]{90}{n\_clust}}}
      &{<50}&{0.39}&{0.46}&{0.39}&{0.18}&{0.33}&{0.23}&{0.27}&{0.40}&{0.35}\\
    {}&{50-70}&{0.33}&{0.37}&{0.34}&{0.14}&{0.24}&{0.18}&{0.26}&{0.31}&{0.32}\\
    {}&{>70}&{0.33}&{0.35}&{0.27}&{0.14}&{0.2}&{0.13}&{0.24}&{0.27}&{0.24}\\
\hline
\end{tabular}
\caption{Averaged Kendall's $\tau$ correlation between the IUN score (by LLM, B or D) and {$D^{90}_{50}$} of the clusters from clustering with embeddings MPN, L6 or E5, UMAPed to dim=20. 
The average is taken on splits by dataset (top 4 rows), by data size (next 4 rows), by clustering algorithm (next 3 rows), and by number of clusters (last 3 rows). Spearman correlations are given in Appendix ~\ref{apx:spearman}.}
\label{tab:corrs_Perce90_50_UMAP1}
\end{table*}

\begin{table*}[th!]
\centering
\begin{tabular}{@{}lllllllllll@{}}
\toprule
\multicolumn{2}{r}{} & \multicolumn{3}{c}{LLM} & \multicolumn{3}{c}{B} &
\multicolumn{3}{c}{D}\\
{}&{selection}&{MPN}&{L6}&{E5}&{MPN}&{L6}&{E5}&{MPN}&{L6}&{E5}\\
\hline
    \parbox[t]{2mm}{\multirow{4}{*}{\rotatebox[origin=c]{90}{dataset}}}
      &{XS}&{0.10}&{0.10}&{0.06}&{0.11}&{0.11}&{0.09}&{0.08}&{0.08}&{0.06}\\
    {}&{CNN}&{0.10}&{0.23}&{0.16}&{0.08}&{0.17}&{0.18}&{0.08}&{0.20}&{0.15}\\
    {}&{DM}&{0.12}&{0.09}&{0.10}&{0.08}&{0.08}&{0.08}&{0.08}&{0.08}&{0.08}\\
    {}&{WN}&{0.12}&{0.13}&{0.16}&{0.09}&{0.10}&{0.14}&{0.11}&{0.08}&{0.15}\\
\hline
    \parbox[t]{2mm}{\multirow{4}{*}{\rotatebox[origin=c]{90}{data size}}}
      &{5000}&{0.14}&{0.16}&{0.18}&{0.12}&{0.15}&{0.19}&{0.17}&{0.14}&{0.17}\\
    {}&{10000}&{0.14}&{0.19}&{0.19}&{0.12}&{0.19}&{0.21}&{0.14}&{0.18}&{0.19}\\
    {}&{15000}&{0.12}&{0.19}&{0.18}&{0.13}&{0.18}&{0.22}&{0.15}&{0.15}&{0.20}\\
    {}&{20000}&{0.13}&{0.20}&{0.21}&{0.15}&{0.20}&{0.23}&{0.13}&{0.18}&{0.19}\\
\hline
    \parbox[t]{2mm}{\multirow{3}{*}{\rotatebox[origin=c]{90}{clusters}}}
      &{HDBSCAN}&{0.12}&{0.11}&{0.16}&{0.15}&{0.14}&{0.17}&{0.14}&{0.14}&{0.16}\\
    {}&{KMeans}&{0.11}&{0.21}&{0.20}&{0.11}&{0.20}&{0.22}&{0.15}&{0.18}&{0.20}\\
    {}&{Agglomerative}&{0.13}&{0.19}&{0.20}&{0.12}&{0.19}&{0.22}&{0.15}&{0.16}&{0.20}\\
\hline
    \parbox[t]{2mm}{\multirow{3}{*}{\rotatebox[origin=c]{90}{n\_clust}}}
      &{<50}&{0.17}&{0.14}&{0.18}&{0.17}&{0.16}&{0.21}&{0.17}&{0.13}&{0.18}\\
    {}&{50-70}&{0.11}&{0.19}&{0.18}&{0.12}&{0.18}&{0.21}&{0.15}&{0.18}&{0.18}\\
    {}&{>70}&{0.09}&{0.20}&{0.19}&{0.09}&{0.17}&{0.21}&{0.13}&{0.16}&{0.19}\\
\hline
\end{tabular}
\caption{Standard deviation of Kendall's $\tau$ correlations between IUN and {$D^{90}_{50}$}. The averages of the correlations are in Table ~\ref{tab:corrs_Perce90_50_UMAP1}.}
\label{tab:corrs_dev_Perce90_50_UMAP1}
\end{table*}

Our approach to estimating IUN from a clustering result is based on an assumption that important urgent news may appear, at least to some extent, in coverage of multiple topics. This would make some clusters closer than they would be otherwise. From the point of view of a cluster with high IUN score, the clusters that are not too far would be 'pulled' closer to the cluster.
In exploring this assumption, we observed that simple features reflecting such 'pull' turn out to correlate well with LLM-generated IUN. Here we present an example of such simple cluster feature, the difference between two percentiles of the distances:
\begin{equation}
\label{eq:D90-D50}
D^{90}_{50} = D_{90} - D_{50}
\end{equation}
To calculate $D_{p}$ for the cluster, $p$ being any number between $0$ and $100$, we gather distances from the center of the cluster to the centers of all the other clusters. Then $D_{p}$ is a distance corresponding to the percentile $p$ of these distances. Essentially, $D^{90}_{50}$ is a difference between the 'radius' of the data (from the point of view of the cluster) and the median distance. The 'radius' is taken at $90\%$ rather than $100\%$ to make it robust against outlier clusters. The median is a robust measure of the 'pull' of some clusters by importance and novelty of the cluster.

\begin{table*}[th!]
\centering
\begin{tabular}{@{}lllllllllll@{}}
\toprule
\multicolumn{2}{r}{} & \multicolumn{3}{c}{LLM} & \multicolumn{3}{c}{B} &
\multicolumn{3}{c}{D}\\
{}&{selection}&{MPN}&{L6}&{E5}&{MPN}&{L6}&{E5}&{MPN}&{L6}&{E5}\\
\hline
    \parbox[t]{2mm}{\multirow{4}{*}{\rotatebox[origin=c]{90}{dataset}}}
      &{XS}&{1.00}&{1.00}&{1.00}&{1.00}&{0.99}&{1.00}&{1.00}&{1.00}&{1.00}\\
    {}&{CNN}&{1.00}&{0.87}&{0.75}&{0.79}&{0.61}&{0.36}&{0.99}&{0.79}&{0.76}\\
    {}&{DM}&{1.00}&{1.00}&{0.99}&{0.82}&{1.00}&{0.95}&{0.80}&{1.00}&{0.98}\\
    {}&{WN}&{1.00}&{0.99}&{0.99}&{1.00}&{1.00}&{0.98}&{0.99}&{1.00}&{0.99}\\
\hline
    \parbox[t]{2mm}{\multirow{4}{*}{\rotatebox[origin=c]{90}{data size}}}
      &{5000}&{1.00}&{0.98}&{0.95}&{0.92}&{0.94}&{0.81}&{0.97}&{0.98}&{0.95}\\
    {}&{10000}&{1.00}&{0.96}&{0.92}&{0.89}&{0.89}&{0.83}&{0.93}&{0.93}&{0.92}\\
    {}&{15000}&{1.00}&{0.97}&{0.93}&{0.92}&{0.91}&{0.85}&{0.94}&{0.94}&{0.93}\\
    {}&{20000}&{1.00}&{0.94}&{0.92}&{0.89}&{0.85}&{0.79}&{0.94}&{0.94}&{0.92}\\
\hline
    \parbox[t]{2mm}{\multirow{3}{*}{\rotatebox[origin=c]{90}{clusters}}}
      &{HDBSCAN}&{1.00}&{0.99}&{0.99}&{0.94}&{0.96}&{0.95}&{0.99}&{0.99}&{0.99}\\
    {}&{KMeans}&{1.00}&{0.94}&{0.92}&{0.88}&{0.89}&{0.74}&{0.93}&{0.91}&{0.92}\\
    {}&{Agglomerative}&{1.00}&{0.97}&{0.88}&{0.89}&{0.85}&{0.78}&{0.92}&{0.94}&{0.90}\\
\hline
    \parbox[t]{2mm}{\multirow{3}{*}{\rotatebox[origin=c]{90}{n\_clust}}}
      &{<50}&{1.00}&{0.99}&{0.99}&{0.85}&{0.99}&{0.85}&{0.94}&{1.00}&{0.99}\\
    {}&{50-70}&{1.00}&{0.96}&{0.96}&{0.89}&{0.87}&{0.83}&{0.94}&{0.93}&{0.96}\\
    {}&{>70}&{1.00}&{0.94}&{0.85}&{0.97}&{0.83}&{0.78}&{0.96}&{0.92}&{0.84}\\
\hline
\end{tabular}
\caption{Fraction of positive Kendall's $\tau$ correlations between IUN and {$D^{90}_{50}$}. The averages of the correlations are in Table ~\ref{tab:corrs_Perce90_50_UMAP1}.}
\label{tab:corrs_fPositive_Perce90_50_UMAP1}
\end{table*}

Correlations between $D^{90}_{50}$ and IUN are shown in Table ~\ref{tab:corrs_Perce90_50_UMAP1}; the correlations are averaged over the considered clustering cases (the number of the cases are in Appendix ~\ref{apx:ncases}).
The clustering cases are split in Table ~\ref{tab:corrs_Perce90_50_UMAP1} in several ways: by dataset, by data size (first N samples are selected from a dataset), by clustering algorithm and by the number of resulting clusters. See Appendix ~\ref{apx:other_umap} for UMAP versions providing correlations similar to UMAP dim=20.

The standard deviation over the clustering cases is in Table ~\ref{tab:corrs_dev_Perce90_50_UMAP1}. In some clustering cases the correlation may happen to be negative (see Table ~\ref{tab:corrs_fPositive_Perce90_50_UMAP1}), but with $MPN$ (all-mpnet-base-v2) embeddings the correlation between $D^{90}_{50}$ and IUN is positive in all the covered clustering cases. 

A more detailed presentation of the accuracy of approximating {\it{IUN}} by $D^{90}_{50}$ is in Appendix ~\ref{apx:curve_distance_ranks}.

In Table ~\ref{tab:features_simple} we attempt to show the role of the 'radius' of the whole dataset $D_{90}$ (from the point of view of the cluster) and the role of the pulled inward 'boundary' $D_{50}$ of the closer half of the dataset (again from the point of view of the cluster). The table is based on all the clustering cases with embedding '{\it{MPN}}' (UMAPed to dimension 20); it shows separately the correlations of $D_{90}$ and $D_{50}$ with LLM-generated IUN score.

\begin{table}[th!]
\centering
\begin{tabular}{@{}cccc@{}}
\hline
{correlation with IUN}&{$avg$}&{$stdev$}&{$F_{pos}$}\\
\hline
    {D90}&{0.07}&{0.15}&{0.7523}\\
    {-D50}&{0.19}&{0.17}&{0.8528}\\
    {D90 - D50}&{0.35}&{0.13}&{1.0000}\\
\hline
\end{tabular}
\caption{Includes separate correlations of the percentiles $D_{90}$ and $-D_{50}$ with IUN. The average ($avg$), standard deviation ($stdev$) and the fraction $F_{pos}$ of positive correlations are taken over all clustering cases with embedding MPN and UMAP dim=20.}
\label{tab:features_simple}
\end{table}

In our measure $D^{90}_{50}$ the 'radius' $D_{90}$ is supposed to be a 'beacon' against which we measure the pull of the boundary $D_{50}$ by the cluster, the pull is stronger if the cluster contains important urgent news. Actually, from the $D_{90}$ row of the Table ~\ref{tab:features_simple} we see that the averaged correlation is positive not only for $-D_{50}$, but (to much less extend) for $D_{90}$. We speculate that a cluster with important urgent news may appear, due to its novelty, at a location a bit further from established unrelated news. 
As expected, $D_{90}$ and $-D_{50}$ combined deliver much better correlation with IUN than if considered separately. In Appendix ~\ref{apx:other_features} we show a few other cluster properties, some of which may be as good or almost as good as $D^{90}_{50}$ (e.g. a replacement of $D_{50}$ by a percentile within $15\%-55\%$), and some could be even better (e.g. $2D_{90} - D_{50} - D_{20}$). 

\section{Conclusion}
\label{sec:conclusion}
We have presented strong correlation between a simple property $D^{90}_{50}$ of clusters in a clustered dataset and the LLM scored IUN (importance and urgency of news). We provided an interpretation of the correlation, and found the correlation to be robust across several news datasets, data sizes, clustering algorithms and text embeddings. We suggest that using this or similar property should help in a speedy finding of important urgent news, filtering unimportant news and ranking clusters by IUN.

\section{Limitations}
\label{sec:Limitations}
The present work cconsidered datasets consisting of $5K$ - $20K$ texts. Much larger datasets may require adjusting such a simple measure as $D^{90}_{50}$. 

We focused on news only; however different in style, the datasets we used contain news articles. It is interesting whether our conclusions would hold for social media or other kinds of texts that may contain news.

\section*{Acknowledgements}
We thank Randy Sawaya for many discussions and review of the paper.

\bibliography{HowDiscernIUN}

\begin{thebibliography}{17}
\expandafter\ifx\csname natexlab\endcsname\relax\def\natexlab#1{#1}\fi

\bibitem[{Campello et~al.(2013)Campello, Moulavi, and Sander}]{10.1007/978-3-642-37456-2-14}
Ricardo J. G.~B. Campello, Davoud Moulavi, and Joerg Sander. 2013.
\newblock Density-based clustering based on hierarchical density estimates.
\newblock In \emph{Advances in Knowledge Discovery and Data Mining}, pages 160--172, Berlin, Heidelberg. Springer Berlin Heidelberg.

\bibitem[{Fatemi and Hu(2023)}]{fatemi2023comparative}
Sorouralsadat Fatemi and Yuheng Hu. 2023.
\newblock \href {https://arxiv.org/abs/2312.08725} {A comparative analysis of fine-tuned {LLM}s and few-shot learning of {LLM}s for financial sentiment analysis}.
\newblock \emph{arXiv}, arXiv:2312.08725.

\bibitem[{Gilardi et~al.(2023)Gilardi, Alizadeh, and Kubli}]{Gilardi2023}
Fabrizio Gilardi, Meysam Alizadeh, and Maël Kubli. 2023.
\newblock \href {https://doi.org/10.1073/pnas.2305016120} {Chat{GPT} outperforms crowd workers for text-annotation tasks}.
\newblock \emph{Proceedings of the National Academy of Sciences}, 120(30).

\bibitem[{Hermann et~al.(2015)Hermann, Kocisky, Grefenstette, Espeholt, Kay, Suleyman, and Blunsom}]{cnndm}
Karl~Moritz Hermann, Tomas Kocisky, Edward Grefenstette, Lasse Espeholt, Will Kay, Mustafa Suleyman, and Phil Blunsom. 2015.
\newblock \href {https://proceedings.neurips.cc/paper/2015/file/afdec7005cc9f14302cd0474fd0f3c96-Paper.pdf} {Teaching machines to read and comprehend}.
\newblock In \emph{Advances in Neural Information Processing Systems}, volume~28, pages 1693--1701. Curran Associates, Inc.

\bibitem[{Kuzman et~al.(2023)Kuzman, Mozetič, and Ljubešić}]{kuzman2023chatgpt}
Taja Kuzman, Igor Mozetič, and Nikola Ljubešić. 2023.
\newblock \href {https://arxiv.org/abs/2303.03953} {Chat{GPT}: Beginning of an end of manual linguistic data annotation? {U}se case of automatic genre identification}.
\newblock \emph{arXiv}, arXiv:2303.03953.

\bibitem[{Lu et~al.(2023)Lu, Yao, Zhang, Wang, Zhang, Lu, Li, and Wang}]{lu2023human}
Yuxuan Lu, Bingsheng Yao, Shao Zhang, Yun Wang, Peng Zhang, Tun Lu, Toby Jia-Jun Li, and Dakuo Wang. 2023.
\newblock \href {https://arxiv.org/abs/2311.09825} {Human still wins over {LLM}: An empirical study of active learning on domain-specific annotation tasks}.
\newblock \emph{arXiv}, arXiv:2311.09825.

\bibitem[{McInnes and Healy(2017)}]{8215642}
Leland McInnes and John Healy. 2017.
\newblock \href {https://doi.org/10.1109/ICDMW.2017.12} {Accelerated hierarchical density based clustering}.
\newblock In \emph{2017 IEEE International Conference on Data Mining Workshops (ICDMW)}, pages 33--42.

\bibitem[{McInnes et~al.(2020)McInnes, Healy, and Melville}]{mcinnes2020umap}
Leland McInnes, John Healy, and James Melville. 2020.
\newblock \href {https://arxiv.org/abs/1802.03426} {{UMAP}: Uniform manifold approximation and projection for dimension reduction}.
\newblock \emph{arXiv}, arXiv:1802.03426.

\bibitem[{Narayan et~al.(2018)Narayan, Cohen, and Lapata}]{narayan-etal-2018-dont}
Shashi Narayan, Shay~B. Cohen, and Mirella Lapata. 2018.
\newblock \href {https://doi.org/10.18653/v1/D18-1206} {Don{'}t give me the details, just the summary! {T}opic-aware convolutional neural networks for extreme summarization}.
\newblock In \emph{Proceedings of the 2018 Conference on Empirical Methods in Natural Language Processing}, pages 1797--1807, Brussels, Belgium. Association for Computational Linguistics.

\bibitem[{Reimers and Gurevych(2019)}]{reimers-gurevych-2019-sentence}
Nils Reimers and Iryna Gurevych. 2019.
\newblock \href {https://doi.org/10.18653/v1/D19-1410} {Sentence-{BERT}: Sentence embeddings using {S}iamese {BERT}-networks}.
\newblock In \emph{Proceedings of the 2019 Conference on Empirical Methods in Natural Language Processing and the 9th International Joint Conference on Natural Language Processing (EMNLP-IJCNLP)}, pages 3982--3992, Hong Kong, China. Association for Computational Linguistics.

\bibitem[{See et~al.(2017)See, Liu, and Manning}]{see-etal-2017-get}
Abigail See, Peter~J. Liu, and Christopher~D. Manning. 2017.
\newblock \href {https://doi.org/10.18653/v1/P17-1099} {Get to the point: Summarization with pointer-generator networks}.
\newblock In \emph{Proceedings of the 55th Annual Meeting of the Association for Computational Linguistics (Volume 1: Long Papers)}, pages 1073--1083, Vancouver, Canada. Association for Computational Linguistics.

\bibitem[{Törnberg(2023)}]{Petter2023chatgpt4}
Petter Törnberg. 2023.
\newblock \href {https://arxiv.org/abs/2304.06588} {{ChatGPT}-4 outperforms experts and crowd workers in annotating political twitter messages with zero-shot learning}.
\newblock \emph{arXiv}, arXiv:2304.06588.

\bibitem[{Vasilyev et~al.(2023)Vasilyev, Isono, and Bohannon}]{vasilyev2023linear}
Oleg Vasilyev, Fumika Isono, and John Bohannon. 2023.
\newblock \href {https://arxiv.org/abs/2305.14256} {Linear cross-lingual mapping of sentence embeddings}.
\newblock \emph{arXiv}, arXiv:2305.14256.

\bibitem[{Wang et~al.(2022)Wang, Yang, Huang, Jiao, Yang, Jiang, Majumder, and Wei}]{wang2022text}
Liang Wang, Nan Yang, Xiaolong Huang, Binxing Jiao, Linjun Yang, Daxin Jiang, Rangan Majumder, and Furu Wei. 2022.
\newblock \href {https://arxiv.org/abs/2212.03533} {Text embeddings by weakly-supervised contrastive pre-training}.
\newblock \emph{arXiv}, arXiv:2212.03533.

\bibitem[{Ward(1963)}]{doi:10.1080/01621459.1963.10500845}
Joe~H. Ward. 1963.
\newblock \href {https://doi.org/10.1080/01621459.1963.10500845} {Hierarchical grouping to optimize an objective function}.
\newblock \emph{Journal of the American Statistical Association}, 58(301):236--244.

\bibitem[{Weber and Reichardt(2023)}]{weber2023evaluation}
Maximilian Weber and Merle Reichardt. 2023.
\newblock \href {https://arxiv.org/abs/2401.00284} {Evaluation is all you need. prompting generative large language models for annotation tasks in the social sciences. a primer using open models}.
\newblock \emph{arXiv}, arXiv:2401.00284.

\bibitem[{Yin et~al.(2019)Yin, Hay, and Roth}]{yin-etal-2019-benchmarking}
Wenpeng Yin, Jamaal Hay, and Dan Roth. 2019.
\newblock \href {https://doi.org/10.18653/v1/D19-1404} {Benchmarking zero-shot text classification: Datasets, evaluation and entailment approach}.
\newblock In \emph{Proceedings of the 2019 Conference on Empirical Methods in Natural Language Processing and the 9th International Joint Conference on Natural Language Processing (EMNLP-IJCNLP)}, pages 3914--3923, Hong Kong, China. Association for Computational Linguistics.

\end{thebibliography}

\appendix

\section{Scoring IUN}
\label{apx:scoring}
\subsection{Prompt for IUN}
\label{apx:ssec:prompt}

The prompt used for scoring IUN (see Section ~\ref{ssec:data_scoring}) is shown in Figure ~\ref{fig:prompt}. The GPT-3.5-Turbo was used with temperature set to zero and maximal number of output tokens set to 1. Any result that is not one of the tokens $"1", "2", "3", "4", "5"$ was considered as a scoring failure and was excluded from the dataset. The fraction of the failures was below $1\%$. Allowing longer output indicates that many failures come from the GPT trying to output not a single token but a sentence, either describing or explaining the result. 

\begin{figure}[th!]
\hrulefill\par
\vspace{3pt}
\small{
Assign a score in Likert scale (1 to 5) to rate the importance and urgency of a news article.
          
Your answer should contain only one digit: 1, 2, 3, 4 or 5.
          
Here is a simple guide for assigning the score to the text:

\hspace{10pt}1: The text is not a news article.

\hspace{10pt}2: The news in the text can be perceived as not important and not urgent.

\hspace{10pt}3: The news in the text can be perceived as having low importance and low urgency.

\hspace{10pt}4: The news in the text can be perceived as important and urgent.

\hspace{10pt}5: The news in the text can be perceived as highly important and urgent.

This is the text:
}

\hrulefill\par
\caption{Part of the prompt preceding the text. Used for scoring IUN.}
\label{fig:prompt}
\end{figure}

The frequency of failures can be somewhat decreased by adding repeated attempts to score, with gradual increase of the temperature and with selection of the first non-failed result. Since the fraction of the failures is less than $1\%$ even at zero temperature, for simplicity and easy repeatability we used only the 'immediately correct' zero-temperature scores.

We have also tried several other versions of the prompts. Example of a system-user prompt version is in Figure ~\ref{fig:prompt-system}. Good prompts can decrease the fraction of failures, but we have never achieved the scoring completely free of failures across all four news datasets used in this work. Difference in scoring results provided by prompting is negligible if the definitions of the Likert scale are kept the same, and if the statements about the scoring goal are clearly separated. However, using complicated sentences or not insisting clearly enough on the format of the output (single token between "1" and "5") leads to strong increase of the scoring failures.

\begin{figure}[th!]
\hrulefill\par
\vspace{3pt}
\small{
You are a labeler, skilled in rating the importance and urgency of news.

You are using Likert scale (1 to 5):

\hspace{10pt}1: The text does not have news.

\hspace{10pt}2: The news in the text can be perceived as not important and not urgent.

\hspace{10pt}3: The news in the text can be perceived as having low importance and low urgency.

\hspace{10pt}4: The news in the text can be perceived as important and urgent.

\hspace{10pt}5: The news in the text can be perceived as highly important and urgent.

Rate the provided text. Respond with one digit: 1, 2, 3, 4 or 5.
}

\hrulefill\par
\caption{Version of the IUN scoring prompt as a system message.}
\label{fig:prompt-system}
\end{figure}

\subsection{Classification by light models}
\label{apx:ssec:classif}
A faster alternative to LLM scoring is by classification using light models. In Section ~\ref{ssec:data_scoring} we used classes ['urgent', 'not urgent']. The resulting score turned out correlate well with the LLM score (Table ~\ref{ssec:data_scoring}). Other choices, such as ['important', 'not important'], ['news', 'not news'] have much weaker correlations with the LLM generated score, and provided less convincing examples of scores. We conclude that the urgency may be the main indicator of truly "important urgent news".

\subsection{Examples of scored articles}
\label{apx:ssec:text_examples}
To demonstrate LLM-generated IUN on several examples, for each IUN score from $5$ down to $1$ we select the very first couple of articles in dataset XSum (BBC news) for which LLM (GPT3.5-Turbo) produced that score. The score is taken for the top chunk of the text, but for easier review of our examples, in Figure ~\ref{fig:text_examples_XS} we show the summaries of the corresponding texts.

\begin{figure}[th!]
\hrulefill\par
\small{
{\bf{5}}\hspace{10pt}
An extensive aerial and underwater survey has revealed that 93\% of Australia's Great Barrier Reef has been affected by coral bleaching.

{\bf{5}}\hspace{10pt}
The Queen has made her first visit to a World War Two concentration camp, Bergen-Belsen, in northern Germany.

{\bf{4}}\hspace{10pt}
Clean-up operations are continuing across the Scottish Borders and Dumfries and Galloway after flooding caused by Storm Frank.

{\bf{4}}\hspace{10pt}
Two tourist buses have been destroyed by fire in a suspected arson attack in Belfast city centre.

{\bf{3}}\hspace{10pt}
Lewis Hamilton stormed to pole position at the Bahrain Grand Prix ahead of Mercedes team-mate Nico Rosberg.

{\bf{3}}\hspace{10pt}
Manchester City midfielder Ilkay Gundogan says it has been mentally tough to overcome a third major injury.

{\bf{2}}\hspace{10pt}
Defending Pro12 champions Glasgow Warriors bagged a late bonus-point victory over the Dragons despite a host of absentees and two yellow cards.

{\bf{2}}\hspace{10pt}
Newport Gwent Dragons number eight Ed Jackson has undergone shoulder surgery and faces a spell on the sidelines.

{\bf{1}}\hspace{10pt}
Read match reports for Tuesday's 10 games in the Championship, including Newcastle's 6-0 pummelling of Queens Park Rangers.

{\bf{1}}\hspace{10pt}
Double Rio Olympics gold medallist Laura Kenny (nee Trott) recognises the importance of sporting volunteers - the Unsung Heroes - and wants you to nominate yours.

}
\hrulefill\par
\caption{Examples of IUN score generated on first text chunk, and the corresponding summary of the article. The articles are from XSum dataset.}
\label{fig:text_examples_XS}
\end{figure}

Figure ~\ref{fig:text_examples_CNN} does the same with examples from CNN part of CNN/DailyMail dataset.

\begin{figure}[th!]
\hrulefill\par
\small{
{\bf{5}}\hspace{10pt}
NEW: Guinean government says most victims were crushed in the crowd. United Nations, citing media reports, said at least 58 people died. African Union expressed its "grave concern" about the situation.

{\bf{5}}\hspace{10pt}
33 killed in suicide bombing at reconciliation conference in Baghdad. Tuesday's attack came as tribal leaders were attending conference. Bombing came 3 days after Iraqi PM urged nation's sheikhs to join government.

{\bf{4}}\hspace{10pt}
U.S.-based scientists say their data points toward the existence of the Higgs boson. Finding the Higgs boson would help explain the origin of mass. But the research at the Tevatron collider doesn't provide a conclusive answer. Attention now turns to a seminar Wednesday on data from the Large Hadron Collider.

{\bf{4}}\hspace{10pt}
Zimmerman posts \$5,000 bail; he was accused of throwing a bottle at a girlfriend. "He hasn't been very lucky with the ladies," attorney says of Zimmerman. He became a national figure after being charged, then acquitted in Trayvon Martin's death.

{\bf{3}}\hspace{10pt}
London choir is made up of sufferers of neurological conditions, friends and carers. Growing evidence that music has neurological, physical, psychological benefits. Music used to boost rehabilitation of stroke patients, improve motor function. New approaches to music therapy could bring field into mainstream rehab practice.

{\bf{3}}\hspace{10pt}
Barack Obama and George Obama share a father, the late Barack Obama Sr. George Obama denies media reports that he's living on a dollar a day. "I think I wanted to learn about my father the same way he did," George says.

{\bf{2}}\hspace{10pt}
Politics is often a family business -- not exactly what the founding fathers intended. Nonetheless, our country has a long history of political dynasties The Kennedy, Bush, and Clinton families are just a few of the political dynasties. The Dingells have been in Congress since the Great Depression.

{\bf{2}}\hspace{10pt}
Rickie Fowler unveils a patriotic haircut ahead of this week's Ryder Cup. The American has "USA" shaved into the side of his head. The Ryder Cup pits American and European golfers against each other. The 2014 match takes place in Scotland later this week.

{\bf{1}}\hspace{10pt}
Emmerich developed obsession with the weather during filming of "10,000 BC". Film was shot in New Zealand's South Island, South Africa and Namibia. Other challenges include creating film's 'terror birds', shark-like predators. Miniature pyramids, 'God's palace', made in Munich then shipped to Namibia.

{\bf{1}}\hspace{10pt}
Perveen Crawford, Hong Kong's first female pilot, shows us around her favorite spots. For the best seafood try Po Toi O a small fishing village in the New Territories. The retro-chic China Club in Central Hong Kong serves traditional Chinese food.

}
\hrulefill\par
\caption{Examples of IUN score generated on first text chunk, and the corresponding summary of the article. The articles are from CNN part of CNN/DailyMail dataset.}
\label{fig:text_examples_CNN}
\end{figure}

\section{Clustering}
\label{apx:clustering_algos}
In Section ~\ref{ssec:data_clustering} we described the clustering data used in this work. Here are the parameters used with clustering algorithms:

{\it{HDBSCAN-flat}}: min cluster size $5$; cluster selection method '{\it{eom}}'.

{\it{KMeans}}: '{\it{lloyd}}', initialization 'k-means++'; {\it{n\_init}=$1$}; max iterations $300$; fixed random state.

{\it{Agglomerative}}: linkage {\it{ward}}'.

Our choice of clustering algorithms and parameters to present here is motivated by simplicity and by getting generally good clustering results (by intrinsic evaluation measures) for the datasets, data sizes and embeddings that we have considered.

\section{Distribution of IUN Cluster Scores}
\label{apx:histograms_clusters_IUN}
\subsection{Spread of in-cluster IUN scores}
\label{apx:ssec:histo_scores_dev}

\begin{figure}[]
\includegraphics[width=0.48\textwidth]{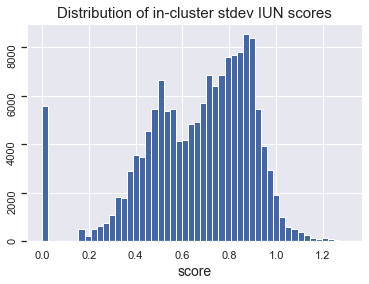}
\caption{Histogram (50 bins) of standard deviations of a cluster's LLM generated IUN scores. The histogram is for the same clusters as Figure ~\ref{fig:Scores_All}.}
\label{fig:Scores_All_stdev}
\end{figure}

In Section ~\ref{ssec:data_clustering} Figure ~\ref{fig:Scores_All} we observed distribution of LLM-generated IUN scores of clusters, where each cluster's score is defined as average over the scores of its items (texts). The distribution is reasonably wide, and this already suggests that IUN scores inside a typical cluster are not distributed randomly but grouped close. Indeed, once we take standard deviation of IUN scores inside each cluster, and make a histogram of these standard deviations, we confirm that typical cluster's standard deviation of IUN is lower than $1.0$: the histogram is shown in Figure ~\ref{fig:Scores_All_stdev}.

\subsection{IUN distribution dependencies}
\label{apx:ssec:histo_scores_split}
For additional illustrations of the distribution of LLM-generated IUN score (Section ~\ref{ssec:data_clustering}, Figure ~\ref{fig:Scores_All}) here we show the distribution on wider data: we include four UMAP versions: UMAP with dimension 20 and number of neighbors 10, UMAP with dimension 10 and number of neighbors 10 or 30, and no UMAP (the original embeddings). We present the distribution counted into bins 1-2, 2-3, 3-4 and 4-5 and split in several ways.

\begin{figure}[]
\includegraphics[width=0.48\textwidth]{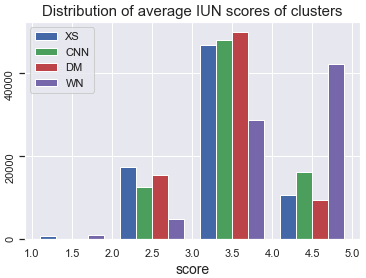}
\caption{4-bins Histogram of LLM generated IUN average cluster scores (bins 1-2, 2-3, 3-4 and 4-5) from all considered clustering cases, split by dataset.}
\label{fig:Scores_Dataset}
\end{figure}

\begin{figure}[]
\includegraphics[width=0.48\textwidth]{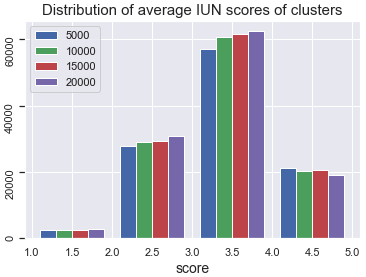}
\caption{Histogram of cluster IUN, as in Figure ~\ref{fig:Scores_Dataset}, but split here by data size.}
\label{fig:Scores_Datasize}
\end{figure}

\begin{figure}[]
\includegraphics[width=0.48\textwidth]{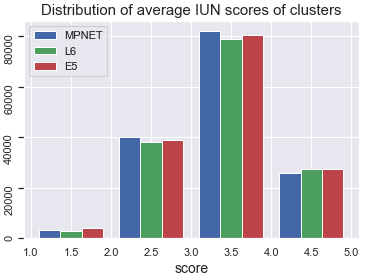}
\caption{Histogram of cluster IUN, as in Figure ~\ref{fig:Scores_Dataset}, but split here by embedding type.}
\label{fig:Scores_Emb}
\end{figure}

In Figure ~\ref{fig:Scores_Dataset} the distribution is split by the datasets. The WN dataset differs from the other datasets, as may be expected given its special generation\footnote{github.com/PrimerAI/primer-research/tree/main/wikinews}.
We speculate that a relatively high (compared to other news datasets) fraction of articles with high IUN in WN is due to the fact that all the articles were in WikiNews\footnote{https://www.wikinews.org/}.

\begin{figure}[]
\includegraphics[width=0.48\textwidth]{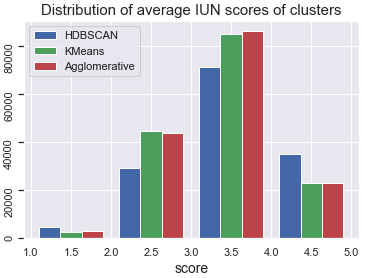}
\caption{Histogram of cluster IUN, as in Figure ~\ref{fig:Scores_Dataset}, but split here by clustering algorithm.}
\label{fig:Scores_Algo}
\end{figure}

\begin{figure}[]
\includegraphics[width=0.48\textwidth]{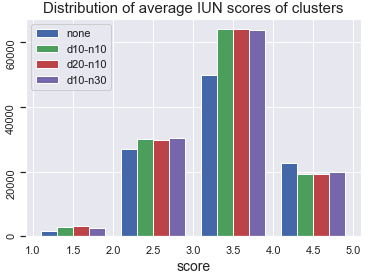}
\caption{Histogram of cluster IUN, as in Figure ~\ref{fig:Scores_Dataset}, but split here by UMAP: 'dN-nM' means a UMAP to dimension N, with number of neighbors M.}
\label{fig:Scores_UMAP}
\end{figure}

Figure ~\ref{fig:Scores_Datasize} shows that, naturally, there is only very weak dependency on the data size. It is encouraging that there is no strong dependency on the embeddings used, Figure ~\ref{fig:Scores_Emb}. However, there is a big difference between the distribution for HDBSCAN vs two other clustering algorithms, Figure ~\ref{fig:Scores_Algo}; in Section ~\ref{sec:observations} we observed that HDBSCAN also differs by providing higher correlation between $D^{90}_{50}$ and IUN. Finally, no-UMAP has a distribution somewhat different from UMAP (Figure ~\ref{fig:Scores_UMAP}). Notice that the total number of cases may be not the same for different splits, because we consider only clustering cases that are successful in producing between 20 and 100 clusters with each cluster containing at least 3 items.

\section{Number of Clustering Cases}
\label{apx:ncases}
This is the number of clustering cases used in Tables ~\ref{tab:corrs_Perce90_50_UMAP1}, ~\ref{tab:corrs_dev_Perce90_50_UMAP1} and ~\ref{tab:corrs_fPositive_Perce90_50_UMAP1} in Section ~\ref{sec:observations}.

\begin{table}[th!]
\centering
\begin{tabular}{@{}llccccc@{}}
\hline
{}&{selection}&{\it{MPN}}&{\it{L6}}&{\it{E5}}\\
\hline
    \parbox[t]{2mm}{\multirow{4}{*}{\rotatebox[origin=c]{90}{dataset}}}
      &{XS}&{107}&{107}&{105}\\
    {}&{CNN}&{107}&{107}&{107}\\
    {}&{DM}&{107}&{103}&{103}\\
    {}&{WN}&{107}&{106}&{107}\\
\hline
    \parbox[t]{2mm}{\multirow{4}{*}{\rotatebox[origin=c]{90}{data size}}}
      &{5000}&{106}&{105}&{105}\\
    {}&{10000}&{107}&{106}&{105}\\
    {}&{15000}&{107}&{104}&{106}\\
    {}&{20000}&{108}&{108}&{106}\\
\hline
    \parbox[t]{2mm}{\multirow{3}{*}{\rotatebox[origin=c]{90}{clusters}}}
      &{HDBSCAN}&{140}&{135}&{134}\\
    {}&{KMeans}&{144}&{144}&{144}\\
    {}&{Agglo}&{144}&{144}&{144}\\
\hline
    \parbox[t]{2mm}{\multirow{3}{*}{\rotatebox[origin=c]{90}{n\_clust}}}
      &{<50}&{144}&{136}&{138}\\
    {}&{50-70}&{140}&{143}&{141}\\
    {}&{>70}&{144}&{144}&{143}\\
\hline
\end{tabular}
\caption{Number of clustering cases}
\label{tab:num_cases_UMAP1}
\end{table}

\begin{table*}[th!]
\centering
\begin{tabular}{@{}lllllllllll@{}}
\toprule
\multicolumn{2}{r}{} & \multicolumn{3}{c}{LLM} & \multicolumn{3}{c}{B} &
\multicolumn{3}{c}{D}\\
{}&{selection}&{MPN}&{L6}&{E5}&{MPN}&{L6}&{E5}&{MPN}&{L6}&{E5}\\
\hline
    \parbox[t]{2mm}{\multirow{4}{*}{\rotatebox[origin=c]{90}{dataset}}}
    &{XS}&{0.62}&{0.75}&{0.61}&{0.40}&{0.63}&{0.46}&{0.54}&{0.65}&{0.63}\\
    {}&{CNN}&{0.48}&{0.35}&{0.17}&{0.08}&{0.10}&{-0.12}&{0.40}&{0.28}&{0.16}\\
    {}&{DM}&{0.38}&{0.58}&{0.44}&{0.10}&{0.39}&{0.19}&{0.10}&{0.38}&{0.30}\\
    {}&{WN}&{0.52}&{0.51}&{0.68}&{0.35}&{0.37}&{0.54}&{0.49}&{0.55}&{0.63}\\
\hline
    \parbox[t]{2mm}{\multirow{4}{*}{\rotatebox[origin=c]{90}{data size}}}
    &{5000}&{0.50}&{0.55}&{0.44}&{0.23}&{0.37}&{0.24}&{0.42}&{0.49}&{0.42}\\
    {}&{10000}&{0.51}&{0.56}&{0.49}&{0.23}&{0.39}&{0.27}&{0.37}&{0.49}&{0.43}\\
    {}&{15000}&{0.49}&{0.54}&{0.48}&{0.24}&{0.38}&{0.30}&{0.38}&{0.46}&{0.46}\\
    {}&{20000}&{0.51}&{0.55}&{0.49}&{0.24}&{0.36}&{0.27}&{0.36}&{0.43}&{0.42}\\
\hline
    \parbox[t]{2mm}{\multirow{3}{*}{\rotatebox[origin=c]{90}{clusters}}}
    &{HDBSCAN}&{0.59}&{0.63}&{0.55}&{0.28}&{0.42}&{0.36}&{0.43}&{0.49}&{0.48}\\
    {}&{KMeans}&{0.46}&{0.51}&{0.45}&{0.21}&{0.35}&{0.23}&{0.36}&{0.45}&{0.41}\\
    {}&{Agglomerative}&{0.46}&{0.51}&{0.43}&{0.21}&{0.35}&{0.22}&{0.36}&{0.46}&{0.41}\\
\hline
    \parbox[t]{2mm}{\multirow{3}{*}{\rotatebox[origin=c]{90}{n\_clust}}}
    &{<50}&{0.55}&{0.63}&{0.55}&{0.26}&{0.47}&{0.34}&{0.40}&{0.57}&{0.49}\\
    {}&{50-70}&{0.48}&{0.52}&{0.48}&{0.21}&{0.35}&{0.27}&{0.38}&{0.44}&{0.46}\\
    {}&{>70}&{0.48}&{0.50}&{0.40}&{0.22}&{0.30}&{0.20}&{0.36}&{0.39}&{0.35}\\
\hline
\end{tabular}
\caption{Averaged Spearman correlation between the IUN score (by LLM, B or D) and {$D^{90}_{50}$} of the clusters from clustering with embeddings MPN, L6 or E5, UMAPed to dim=20.  
The average is taken on splits by dataset (top 4 rows), by data size (next 4 rows), by clustering algorithm (next 3 rows), and by number of clusters (last 3 rows).}
\label{tab:corrsS_Perce90_50_UMAP1}
\end{table*}

\begin{table*}[th!]
\centering
\begin{tabular}{@{}lllllllllll@{}}
\toprule
\multicolumn{2}{r}{} & \multicolumn{3}{c}{LLM} & \multicolumn{3}{c}{B} &
\multicolumn{3}{c}{D}\\
{}&{selection}&{MPN}&{L6}&{E5}&{MPN}&{L6}&{E5}&{MPN}&{L6}&{E5}\\
\hline
    \parbox[t]{2mm}{\multirow{4}{*}{\rotatebox[origin=c]{90}{dataset}}}
    &{SX}&{0.10}&{0.11}&{0.08}&{0.14}&{0.13}&{0.11}&{0.10}&{0.09}&{0.07}\\
    {}&{CNN}&{0.12}&{0.32}&{0.23}&{0.13}&{0.26}&{0.27}&{0.11}&{0.30}&{0.21}\\
    {}&{DM}&{0.16}&{0.11}&{0.13}&{0.12}&{0.11}&{0.11}&{0.13}&{0.11}&{0.13}\\
    {}&{WN}&{0.13}&{0.16}&{0.20}&{0.11}&{0.13}&{0.19}&{0.14}&{0.10}&{0.18}\\
\hline
    \parbox[t]{2mm}{\multirow{4}{*}{\rotatebox[origin=c]{90}{data size}}}
    &{5000}&{0.17}&{0.21}&{0.25}&{0.17}&{0.21}&{0.28}&{0.23}&{0.19}&{0.24}\\
    {}&{10000}&{0.16}&{0.25}&{0.26}&{0.18}&{0.26}&{0.33}&{0.21}&{0.24}&{0.27}\\
    {}&{15000}&{0.15}&{0.25}&{0.24}&{0.19}&{0.25}&{0.32}&{0.21}&{0.21}&{0.26}\\
    {}&{20000}&{0.15}&{0.26}&{0.28}&{0.21}&{0.28}&{0.34}&{0.19}&{0.25}&{0.27}\\
\hline
    \parbox[t]{2mm}{\multirow{3}{*}{\rotatebox[origin=c]{90}{clusters}}}
    &{HDBSCAN}&{0.14}&{0.13}&{0.19}&{0.21}&{0.19}&{0.23}&{0.19}&{0.18}&{0.21}\\
    {}&{KMeans}&{0.14}&{0.29}&{0.27}&{0.18}&{0.29}&{0.35}&{0.21}&{0.25}&{0.28}\\
    {}&{Agglomerative}&{0.15}&{0.26}&{0.28}&{0.17}&{0.27}&{0.34}&{0.22}&{0.23}&{0.28}\\
\hline
    \parbox[t]{2mm}{\multirow{3}{*}{\rotatebox[origin=c]{90}{n\_clust}}}
    &{<50}&{0.20}&{0.16}&{0.23}&{0.24}&{0.21}&{0.30}&{0.23}&{0.15}&{0.23}\\
    {}&{50-70}&{0.14}&{0.25}&{0.25}&{0.18}&{0.27}&{0.32}&{0.21}&{0.25}&{0.25}\\
    {}&{>70}&{0.11}&{0.28}&{0.27}&{0.13}&{0.25}&{0.32}&{0.18}&{0.23}&{0.28}\\
\hline
\end{tabular}
\caption{Standard deviation of Spearman correlations between IUN and {$D^{90}_{50}$}. The averages of the correlations are in Table ~\ref{tab:corrsS_Perce90_50_UMAP1}.}
\label{tab:corrsS_dev_Perce90_50_UMAP1}
\end{table*}

\begin{table*}[th!]
\centering
\begin{tabular}{@{}lllllllllll@{}}
\toprule
\multicolumn{2}{r}{} & \multicolumn{3}{c}{LLM} & \multicolumn{3}{c}{B} &
\multicolumn{3}{c}{D}\\
{}&{selection}&{MPN}&{L6}&{E5}&{MPN}&{L6}&{E5}&{MPN}&{L6}&{E5}\\
\hline
    \parbox[t]{2mm}{\multirow{4}{*}{\rotatebox[origin=c]{90}{dataset}}}
    &{XS}&{1.00}&{1.00}&{1.00}&{1.00}&{0.99}&{1.00}&{1.00}&{1.00}&{1.00}\\
    {}&{CNN}&{1.00}&{0.86}&{0.71}&{0.74}&{0.62}&{0.36}&{0.99}&{0.81}&{0.75}\\
    {}&{DM}&{1.00}&{1.00}&{0.99}&{0.81}&{1.00}&{0.96}&{0.79}&{1.00}&{0.97}\\
    {}&{WN}&{1.00}&{1.00}&{1.00}&{1.00}&{1.00}&{0.97}&{0.99}&{1.00}&{0.98}\\
\hline
    \parbox[t]{2mm}{\multirow{4}{*}{\rotatebox[origin=c]{90}{data size}}}
    &{5000}&{1.00}&{0.98}&{0.94}&{0.89}&{0.95}&{0.81}&{0.96}&{0.98}&{0.94}\\
    {}&{10000}&{1.00}&{0.95}&{0.91}&{0.88}&{0.90}&{0.83}&{0.93}&{0.93}&{0.91}\\
    {}&{15000}&{1.00}&{0.97}&{0.93}&{0.92}&{0.91}&{0.85}&{0.93}&{0.95}&{0.92}\\
    {}&{20000}&{1.00}&{0.95}&{0.91}&{0.87}&{0.84}&{0.80}&{0.94}&{0.94}&{0.92}\\
\hline
    \parbox[t]{2mm}{\multirow{3}{*}{\rotatebox[origin=c]{90}{clusters}}}
    &{HDBSCAN}&{1.00}&{1.00}&{0.99}&{0.93}&{0.96}&{0.94}&{0.99}&{1.00}&{0.98}\\
    {}&{KMeans}&{1.00}&{0.94}&{0.92}&{0.85}&{0.89}&{0.76}&{0.93}&{0.92}&{0.91}\\
    {}&{Agglomerative}&{1.00}&{0.96}&{0.86}&{0.88}&{0.86}&{0.78}&{0.90}&{0.94}&{0.89}\\
\hline
    \parbox[t]{2mm}{\multirow{3}{*}{\rotatebox[origin=c]{90}{n\_clust}}}
    &{<50}&{1.00}&{1.00}&{0.99}&{0.83}&{0.99}&{0.87}&{0.94}&{1.00}&{0.99}\\
    {}&{50-70}&{1.00}&{0.95}&{0.94}&{0.87}&{0.88}&{0.83}&{0.93}&{0.94}&{0.96}\\
    {}&{>70}&{1.00}&{0.94}&{0.85}&{0.96}&{0.84}&{0.77}&{0.96}&{0.92}&{0.83}\\
\hline
\end{tabular}
\caption{Fraction of positive Spearman correlations between IUN and {$D^{90}_{50}$}. The averages of the correlations are in Table ~\ref{tab:corrsS_Perce90_50_UMAP1}.}
\label{tab:corrsS_fPositive_Perce90_50_UMAP1}
\end{table*}

\section{Spearman Correlations}
\label{apx:spearman}
In Section ~\ref{sec:observations}, Table ~\ref{tab:corrs_Perce90_50_UMAP1} we presented Kendall's $\tau$ correlations between IUN score and {$D^{90}_{50}$}. Here we show the corresponding Spearman correlations in Table ~\ref{tab:corrsS_Perce90_50_UMAP1}, structured exactly the same was as Table ~\ref{tab:corrs_Perce90_50_UMAP1}.

Similarly, Table ~\ref{tab:corrsS_dev_Perce90_50_UMAP1} is structured as Table ~\ref{tab:corrs_dev_Perce90_50_UMAP1} but now using Spearman correlations. And Table ~\ref{tab:corrsS_fPositive_Perce90_50_UMAP1} is structured as Table ~\ref{tab:corrs_fPositive_Perce90_50_UMAP1} but using Spearman correlations.

\begin{table*}[th!]
\centering
\begin{tabular}{@{}lllllllllll@{}}
\toprule
\multicolumn{2}{r}{} & \multicolumn{3}{c}{LLM} & \multicolumn{3}{c}{B} &
\multicolumn{3}{c}{D}\\
{}&{UMAP}&{MPN}&{L6}&{E5}&{MPN}&{L6}&{E5}&{MPN}&{L6}&{E5}\\
\hline
    \parbox[t]{2mm}{\multirow{4}{*}{\rotatebox[origin=c]{90}{$avg$}}}
    &{none}&{0.30}&{0.14}&{0.23}&{0.16}&{0.04}&{0.11}&{0.18}&{0.09}&{0.17}\\
    {}&{d10-n10}&{0.33}&{0.40}&{0.32}&{0.15}&{0.26}&{0.18}&{0.24}&{0.33}&{0.30}\\
    {}&{d10-n30}&{0.38}&{0.38}&{0.34}&{0.21}&{0.26}&{0.2}&{0.34}&{0.32}&{0.30}\\
    {}&{d20-n10}&{0.35}&{0.39}&{0.33}&{0.15}&{0.25}&{0.18}&{0.26}&{0.32}&{0.30}\\
    {}&{d20-n30}&{0.37}&{0.35}&{0.35}&{0.21}&{0.23}&{0.21}&{0.33}&{0.29}&{0.32}\\
\hline
    \parbox[t]{2mm}{\multirow{4}{*}{\rotatebox[origin=c]{90}{$stdev$}}}
    &{none}&{0.23}&{0.27}&{0.18}&{0.21}&{0.25}&{0.17}&{0.19}&{0.20}&{0.18}\\
    {}&{d10-n10}&{0.14}&{0.17}&{0.20}&{0.15}&{0.17}&{0.22}&{0.14}&{0.16}&{0.21}\\
    {}&{d10-n30}&{0.14}&{0.21}&{0.21}&{0.13}&{0.20}&{0.23}&{0.16}&{0.21}&{0.21}\\
    {}&{d20-n10}&{0.13}&{0.18}&{0.19}&{0.13}&{0.18}&{0.21}&{0.15}&{0.16}&{0.19}\\
    {}&{d20-n30}&{0.14}&{0.23}&{0.21}&{0.14}&{0.21}&{0.23}&{0.16}&{0.24}&{0.21}\\
\hline
    \parbox[t]{2mm}{\multirow{3}{*}{\rotatebox[origin=c]{90}{$F_{pos}$}}}
    &{none}&{0.83}&{0.73}&{0.87}&{0.76}&{0.63}&{0.78}&{0.80}&{0.68}&{0.81}\\
    {}&{d10-n10}&{0.98}&{0.96}&{0.91}&{0.86}&{0.90}&{0.82}&{0.94}&{0.96}&{0.90}\\
    {}&{d10-n30}&{0.98}&{0.91}&{0.88}&{0.93}&{0.88}&{0.78}&{0.99}&{0.89}&{0.90}\\
    {}&{d20-n10}&{1.00}&{0.96}&{0.93}&{0.90}&{0.90}&{0.82}&{0.95}&{0.95}&{0.93}\\
    {}&{d20-n30}&{0.98}&{0.87}&{0.91}&{0.92}&{0.84}&{0.79}&{0.99}&{0.85}&{0.90}\\
\hline
\end{tabular}
\caption{Kendall's $\tau$ correlation between the IUN score (by LLM, B or D) and {$D^{90}_{50}$} feature of the clusters from clustering with embeddings MPN, L6 or E5. The clustering is done with all considered (Sections ~\ref{sec:data}) datasets, data sizes, clustering algorithms, embeddings and UMAP versions. The results are split by UMAP version (column UMAP). The top horizontal pane shows average $avg$ of the correlations; the middle one shows standard deviation $stdev$, and the last one shows the fraction of positive correlations $F_{pos}$.}
\label{tab:corrs_other_umaps}
\end{table*}

\section{Examples of Other UMAP}
\label{apx:other_umap}
In table ~\ref{tab:corrs_Perce90_50_UMAP1} and ~\ref{tab:corrs_dev_Perce90_50_UMAP1}  of Section ~\ref{sec:observations} we have shown averages and standard deviations of correlations taken over all the considered cases of clustering with UMAP to dimension 20. The fraction of positive correlations was given in Table ~\ref{tab:corrs_fPositive_Perce90_50_UMAP1}.
Here in Table ~\ref{tab:corrs_other_umaps} we extend the clustering cases to several more UMAP examples: 'none' (original embeddings, no UMAP), and, in notations of the table, a few 'dN-nM', meaning UMAP to dimension N, with number of neighbors M (min distance is $0$).
We observe no essential difference in the results, except that not using UMAP makes the correlations with IUN a bit worse. 
We were motivated to consider these and more UMAP versions in vicinity range of dimension and number of neighbors because such UMAPs lead to good clustering quality (accordingly to common intrinsic evaluation measures) on the datasets and the sizes that we used.

\section{Gaps Between IUN and {$\pmb{D^{90}_{50}}$} Ranks}
\label{apx:curve_distance_ranks}
\begin{figure}[]
\includegraphics[width=0.48\textwidth]{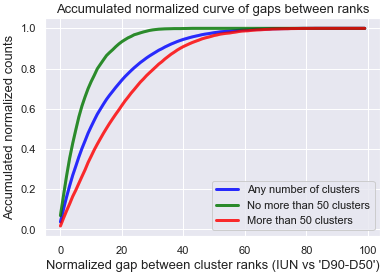}
\caption{Accumulated normalized curve of the differences between the ranks of IUN and {$D^{90}_{50}$}. Axis X shows the difference normalized to 100 by the total number of clusters. Axis Y is the accumulated normalized count of clusters.}
\label{fig:Curve_gaps}
\end{figure}

Kendall's $\tau$ correlation is based on account of concordant and discordant paired observations, and it gives a good account of how agreeable two values (in our case, IUN and {$D^{90}_{50}$}) when ranked. It would be also illustrative to see how far from each other the ranks of the same cluster accordingly to IUN score vs {$D^{90}_{50}$}. For this purpose we show in Figure ~\ref{fig:Curve_gaps} the accumulated normalized curve of the gaps between the ranks. 

The gaps are normalized to total number of clusters in each clustering case. The counts are taken over all our clustering cases with {\it{MPN}} embedding (UMAPed to dim=20). IUN is by LLM (GPT3.5-Turbo). With account of all the clustering cases (blue curve), the gap between the ranks is less than $10\%$ (of the total number of clusters) for half of the clusters. The gap is limited by $50\%$ for $97.9\%$ of the clusters.

In Section ~\ref{sec:observations} we already observed (Tables ~\ref{tab:corrs_Perce90_50_UMAP1}, ~\ref{tab:corrs_dev_Perce90_50_UMAP1}, ~\ref{tab:corrs_fPositive_Perce90_50_UMAP1}) that the relation between IUN and {$D^{90}_{50}$} is stronger for not too high number of clusters. Indeed, accounting for only the clustering cases with more than $50$ clusters (red curve in Figure ~\ref{fig:Curve_gaps}) increases the gaps in the distribution. And limiting the number of clusters to not more than $50$ decreases the gaps (green curve). In this favorable setting, $73.5\%$ of clusters have the gap (between IUN and {$D^{90}_{50}$} ranks) within $10\%$ of the total number of clusters. And $99.2\%$ of clusters have gap within $30\%$, and there are no clusters with gap more than $42\%$.

A few simple measures can help in succinct characterization of the accumulative curve between the ranks. We present such measures as the columns in Table ~\ref{tab:measures_gaps}. The three rows in the table correspond to the three curves of Figure ~\ref{fig:Curve_gaps}. For example, the first row (not more than $51$ clusters) shows that the median of the difference between {$D^{90}_{50}$} and IUN ranks of the same cluster would be $6\%$ of the total number of clusters. This means that, for example, that when averaged over all clustering cases resulting in exactly 50 clusters, the median of the difference between {$D^{90}_{50}$} and IUN ranks will be equal to $3$. The second column shows the average (rather than median) of the difference between the ranks. In the first row it is $7.62\%$ of the total number of clusters. Finally, the third column shows $97.5\%$ percentile $P97.5$. In the first row it shows that for $97.5\%$ of all the clusters the {$D^{90}_{50}$} rank would be withing $26\%$ (of total number of clusters) from IUN rank.

\begin{table}[th!]
\centering
\begin{tabular}{@{}crrr@{}}
\hline
{Number of clusters}&{$median$}&{$avg$}&{$P97.5$}\\
\hline
    {<= 50}&{6}&{7.62}&{26}\\
    {All}&{10}&{14.35}&{49}\\
    {> 50}&{15}&{18.78}&{54}\\
\hline
\end{tabular}
\caption{Median, average ({\it{avg}}) and 97.5\% percentile ({\it{P97.5}}) of gaps between IUN and {$D^{90}_{50}$} ranks. Normalized to 100, similar to axis X in Figure ~\ref{fig:Curve_gaps}.}
\label{tab:measures_gaps}
\end{table}

All this helps in assessing how accurate, in sense of using {$D^{90}_{50}$} instead of IUN, would be a removal of certain number of clusters with lowest {$D^{90}_{50}$} scores, or a selection of certain number of clusters with top {$D^{90}_{50}$} scores. In any case, according to our understanding (discussion of Table ~\ref{tab:features_simple}, Section ~\ref{sec:observations}), the {$D^{90}_{50}$} score reflects importance and urgency of news in its own way, even if it does not exactly coincide with IUN generated by LLM.

\section{Other Examples of Cluster Properties}
\label{apx:other_features}
\begin{table}[th!]
\centering
\begin{tabular}{@{}cccc@{}}
\hline
{correlation with IUN}&{$avg$}&{$stdev$}&{$F_{pos}$}\\
\hline
    {D90-D10}&{0.34}&{0.16}&{0.9930}\\
    {D90-D15}&{0.39}&{0.13}&{0.9953}\\
    {D90-D20}&{0.38}&{0.12}&{0.9977}\\
    {D90-D25}&{0.36}&{0.12}&{0.9977}\\
    {D90-D30}&{0.36}&{0.13}&{0.9977}\\
    {D90-D35}&{0.35}&{0.13}&{0.9953}\\
    {D90-D40}&{0.35}&{0.13}&{0.9977}\\
    {D90-D45}&{0.35}&{0.13}&{0.9977}\\
    {D90-D50}&{0.35}&{0.13}&{1.0000}\\
    {D90-D55}&{0.35}&{0.14}&{0.9977}\\
    {D90-D60}&{0.34}&{0.14}&{0.9883}\\
    {D90-D65}&{0.33}&{0.15}&{0.9860}\\
\hline
\end{tabular}
\caption{Correlation of several similar cluster features (column 1) with IUN. The average ($avg$), standard deviation ($stdev$) and the fraction $F_{pos}$ of positive correlations are taken over all clustering cases with embedding MPN and UMAP dim=20.}
\label{tab:features_more}
\end{table}
In Section ~\ref{sec:observations} we used Table ~\ref{tab:features_simple} for justifying and interpreting the usage of $D_{90} - D_{50}$ as an indicator of IUN. Here in Table ~\ref{tab:features_more} we give examples of a few more similar cluster features, by replacing $D_{50}$ with distance percentiles between $10\%$ and $65\%$. (Changing 'radius' from $90\%$ to $85\%$ or to $95\%$ makes all the rows slightly worse; this is not shown in the table.) 

Using the percentiles $15\%$ - $45\%$ instead of $D_{50}$ provides as good or even better correlations with IUN (at least averaged across all the clustering cases), especially $20\%$. The reason may be in stronger 'pull' in the vicinity of the important urgent news cluster. But it also possible that replacing $D_{50}$ by a lesser percentile, and thus sampling less of the 'pull' volume, makes the correlations less robust (column $F_{pos}$). On the other hand, using a percentile above $50\%$, while adding weaker 'pull' volume, reduces the difference with $D_{90}$. Thus, altogether the $D_{90} - D_{50}$ row in Table ~\ref{tab:features_more} is the safe choice.

In Section ~\ref{sec:observations}, discussing Table ~\ref{tab:features_simple} we considered $D_{90}$ as a 'beacon' (or a 'radius') against which to measure the 'pull' of imporatnce and urgency of the news in the cluster. It is worth to note that ignoring that consideration and choosing the 'beacon' haphazardly leads to worse results. For example, replacing the $90\%$ percentile of the distances by their average '{$A$}' would produce (for {$A - D_{50}$}), in Table ~\ref{tab:features_more}, the values $avg=0.25$, $stdev=0.17$ and $F_{pos}=0.9180$. 

On the other hand, there may be possibilities for measures with even better correlations (with LLM-generated IUN) than {$D_{90} - D_{50}$}. It may be possible to strike a good balance between stronger and weaker pull areas, for example combining $D_{50}$ and $D_{20}$. The measure
\begin{equation}
\label{eq:2D90-D50-D20}
2D_{90} - D_{50} - D_{20}
\end{equation}
would have, in Table ~\ref{tab:features_more}, the values $avg=0.37$, $stdev=0.12$ and $F_{pos}=1.0000$. This is better than $D_{90} - D_{50}$. But simplicity makes always a better promise of robustness, and that is why we left $D^{90}_{50}$ as our recommended example.

\end{document}